\pdfoutput=1
\documentclass[conference]{IEEEtran}
\usepackage[utf8]{inputenc}
\usepackage[T1]{fontenc}
\usepackage{graphicx}
\usepackage{wrapfig}
\usepackage{float}
\usepackage{url}
\usepackage{cite}
\usepackage{algorithmic}
\usepackage{lipsum}
\usepackage{cuted}
\begin{document}
\title{Road Accident Proneness Indicator Based on Time, Weather and Location Specificity Using Graph Neural Networks}

\author{
    Srikanth Chandar\\
    ECE\\
    PES University\\
    srikanth.chandar@gmail.com
  \and
    Anish Reddy\\
    ECE\\
    PES University\\
    anish.reddy652@gmail.com
    \and
    Muvazima Mansoor\\
    ECE\\
    PES University\\
    muvazima99@gmail.com
    \and
    Prof. Suresh Jamadagni\\
    CSE\\
    PES University\\
    sureshjamadagni@pes.edu
}
\date{\today}

\maketitle
\begin{abstract}

In this paper, we present a novel approach to identify the Spatio-temporal and environmental features that influence the safety of a road, and predict its accident proneness based on these features. A total of 14 features were compiled based on Time, Weather and Location (TWL) specificity along a road. To determine the influence each of the 14 features carries, a sensitivity study was performed using Principal Component Analysis. Using the locations of accident warnings, a Safety Index was developed to quantify how accident-prone a particular road is. We implement a novel approach to predict the Safety Index of a road- based on its TWL specificity by using a Graph Neural Network (GNN) architecture. The proposed architecture is uniquely suited for this application due to its ability to capture the complexities of the inherent nonlinear interlinking in a vast feature space. We employed a GNN to emulate the TWL feature vectors as individual nodes which were interlinked vis-à-vis edges of a graph. This model was verified to perform better than Logistic Regression, simple Feed- Forward Neural Networks, and even Long Short Term Memory (LSTM) Neural Networks. We validated our approach on a data set containing the alert locations along the routes of inter-state buses.
The results achieved through this GNN architecture, using a TWL input feature space proved to be more feasible than the other predictive models, having reached a peak accuracy of 65\%. 
\end{abstract}

\section{Introduction}
 Road accidents account for a high fraction of mortality and account for the deaths of approximately 1.35 million people annually [1]. This makes motor vehicle collisions the leading cause of injury and death among children  10–19 years old, worldwide (260,000 children die a year, 10 million are injured) [2]. Middle-income countries have the highest rate with 20 deaths per 100,000 inhabitants, accounting for 80\% of all road fatalities with 52\% of all vehicles [3]. These accidents also result in a substantial amount of non-fatal injuries and property loss. The global economic cost of motor vehicle collisions was estimated at \$518 billion per year in 2003. Traffic collisions affect the national economy as the cost of road injuries are estimated to account for 1.0\% to 2.0\% of the Gross National Product (GNP) of every country each year [4]. Comprehensive studies on accident analysis and prevention thus become imperative. 
 
 Being able to predict how accident-prone a location is, can reduce the risk of calamity. Predictive models based on historical data have been employed to address this issue. However, the majority of existing research on accident prediction is based on human error and behavioral patterns [5,6]. The cause of accidents, however, may be influenced by a wide spectrum of factors. Bollapragada et al.[7] classified the various reasons for the
causes of road accidents and also examined through models about the relationship. A study by Jaroszweski et al [8] showed the effects of adverse weather, on-road hazards, confirming a strong correlation between environmental conditions and accidents.

In this paper, a novel approach using 14 Spatio-temporal factors based on time, weather, and location (TWL) to predict the accident proneness of a road, is presented. The features of the data-set include alarm type, date, time, latitude, longitude, and speed. Spatio-temporal weather and altitude data was gathered using World Weather Online API. Reverse geocoding was then performed to obtain the names of the roads using OpenStreetMaps Nominatim API. Subsequently, a Safety Index was calculated using the spatial concentrations of warnings. A detailed exploratory study and Principal Component Analysis was performed to determine the correlation of the 14 distinct features relating to weather conditions, time, altitude, and speed with the safety index.
 A Graph Neural Network (GNN) was then used to employ a neighborhood aggregation scheme, where the node's representation vector is calculated by recursively aggregating and transforming the representation vectors of its neighboring nodes [9]. This approach gives a better accuracy compared to a simple Feed-Forward network, Logistic Regression, and LSTM.

\section{Methodology}
To predict the accident proneness of a location, a 3 step approach was employed. The data set was first pre-processed and augmented with relevant TWL statistics. Features of weather were compiled using the World Weather API. The data was then normalized with respect to the number of times a particular road had been used. The concentrations of collision warnings in every 1-kilometer radius was calculated using the Haversine formula. Subsequently, a Safety Index was computed to quantify the accident proneness of a location. In the second step, an exploratory study including Principal Component Analysis was conducted to understand the influence of various features on the accident proneness. Finally, multiple predictive models were examined to accurately forecast the Safety Index of a new data entry.

\begin{figure}[H]
\begin{center}
\includegraphics[width=55mm,height=100mm]{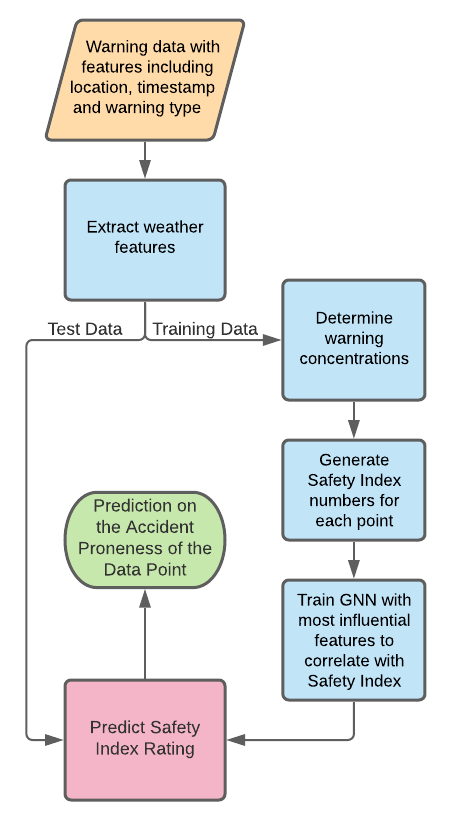} 
\caption{{\bf Process Overview}}
\label{fig1}
\end{center}
\end{figure}

\section{Dataset and Preprocessing}

\subsection{Original Dataset}
The research was initiated with a data set describing features of the journeys of buses along different roads shown below.
\begin{itemize}
\item deviceId: It is used to uniquely identify a bus. The dataset consists of 9 unique buses.
\item alarmType: It indicates the type of alert recorded by the sensor. There are five different types of alerts:
\begin{enumerate}
\item \raggedright HEADWAY MONITORING WARNING (HMW): The headway monitoring warning (HMW) helps drivers maintain a safe following distance from the vehicle ahead of them by providing visual and audible alerts if the distance becomes unsafe.
\item PEDESTRIANS AND CYCLIST DETECTION AND COLLISION WARNING (PCW): The PCW notifies the driver of a pedestrian or cyclist in the danger zone and alerts drivers of an imminent collision with a pedestrian or cyclist.
\item  FORWARD COLLISION WARNINGS (FCW): FCW provides an alert before a possible low-speed collision with the vehicle in front, thus assisting the driver at a low speed in densely heavy traffic.
\item STOPPAGE - stoppage alert is raised when the bus halts. 
\item HIGH BRAKE (HB): High Brake alert is raised when the driver brakes suddenly without slowing down. 
    
\end{enumerate}

\item \raggedright recorded\_at\_hour: the hour of the day at which the alert was recorded.
\item recorded\_at\_date: the date at which the alert was recorded.
\item recorded\_at\_time: the exact time of the day at which the alert was recorded.
\item AM/PM: to determine if the alert was raised before midday or after midday. 
\item latitude: the latitude at which the alert was raised.
\item longitude: the longitude at which the alert was raised
\item speed: speed of the bus at which the alert was raised.
\end{itemize}
\subsection{Feature Extraction}
Using the features present in the original data-set, multiple other features were extracted.
\begin{itemize}
    \item Altitude: the altitude at which the alert was raised was extracted using Digital Elevation Model(DEM) database. 
    \item Address and Name of Road: The address and the road at which the alert was raised was obtained by reverse geocoding the coordinates using OpenStreetMaps Nominatim API. 
    \item \raggedright Weather features: 24 weather features that include- maxtempC, mintempC, sunHour, uvIndex, moon\_illumination, moonrise, moonset, sunrise, sunset, DewPointC, FeelsLikeC, HeatIndexC, WindChillC, WindGustKmph, cloudcover, humidity, precipMM, pressure, tempC, visibility, winddirDegree and windspeedKmph were extracted using the World Weather Online (WWO) API. These features were appended to the original data set and visualized.
\end{itemize}

\begin{figure*}
\begin{center}
\includegraphics[width=180mm,height=80mm]{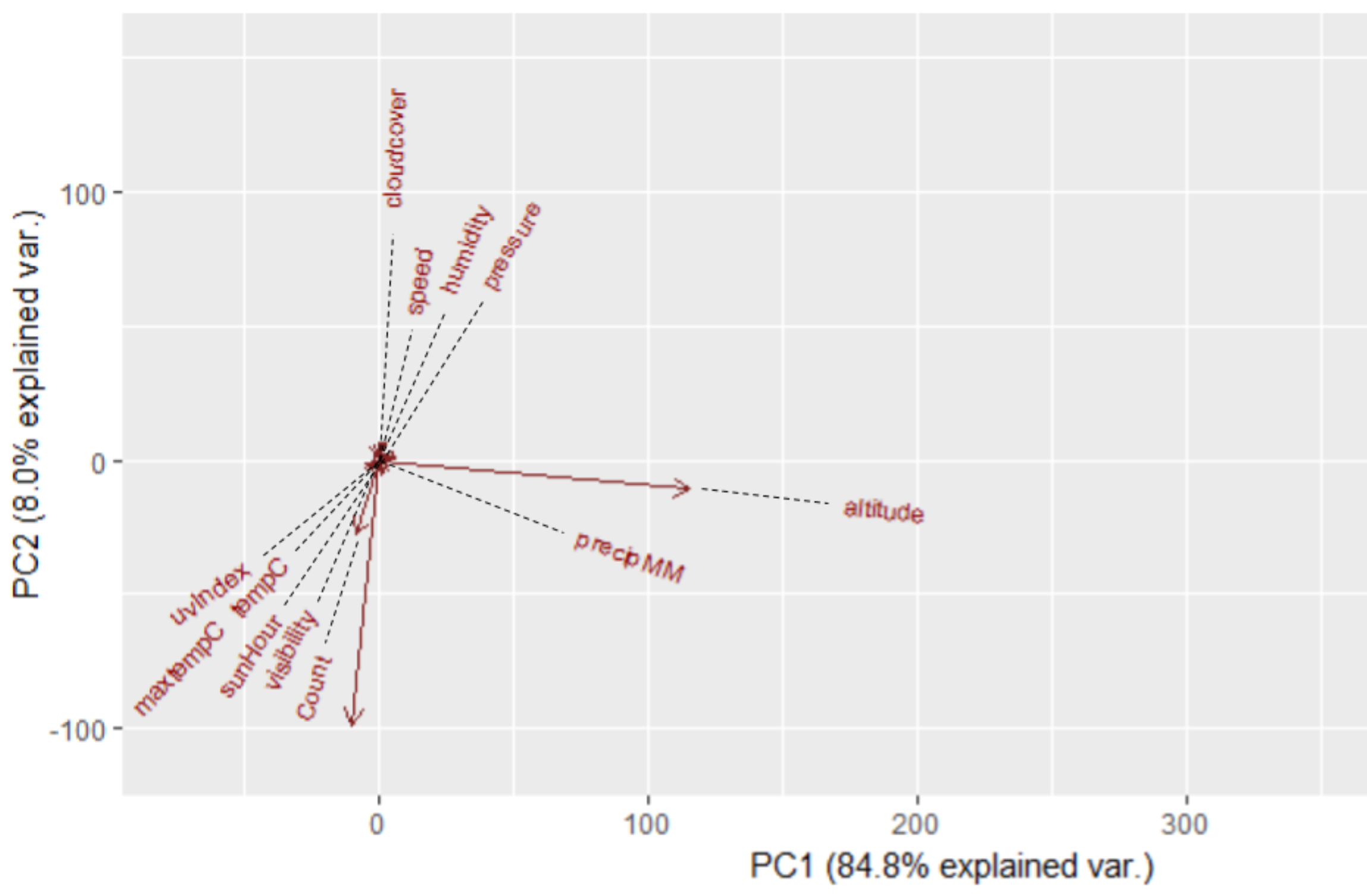} 
\caption{{\bf Principal Component Analysis}}

\label{fig2}
\end{center}
\end{figure*}

\subsection{Safety Index}

The Safety Index is a metric defined to determine the accident proneness of a particular location. For every location, the number of warnings in a one-kilometer radius was calculated using the Haversine formula.

\begin{center}
    $Distance  =$
\end{center}
\begin{equation}{
2r * arcsin({\sqrt {hav(\varphi_{2}-\varphi_{1})+ cos(\varphi _{1})cos(\varphi _{2})hav(\lambda _{2}-\lambda _{1})}})} 
\end{equation}

\bigskip

The number was then normalized based on the number of trips, by the buses, through a particular location, and a non-linear transformation was applied on it. The output of which is an integer from 1 to 5. This integer is the safety index value and assigned to the location as a new feature. If a location has a safety index rating of 1, it implies that there was a relatively lower number of collision warnings, within a 1 km radius of that location. Ergo the area is less accident-prone. Similarly, a rating of 5 implies that a location is more prone to accidents due to a high concentration of collision warnings in its vicinity. The integers from 1 to 5 indicate the level of susceptibility to accidents. 

\section{PCA and Exploratory Data Analysis}
The normalized features of the data set were used to draw useful insights through an exploratory study (Fig.3, Fig.4). Correlation plots between each individual feature and the Safety Index indicated the influence, a feature had on the accident proneness at a particular location. Principal Component Analysis was applied on the feature vectors to calculate the 2 vectors which represent the directions of the highest variance of the data. These vectors were chosen as the axes on a Bi-graph, on which the features were plotted (Fig.2).

\begin{figure}[H]
\begin{center}
\includegraphics[width=85mm,height=50mm]{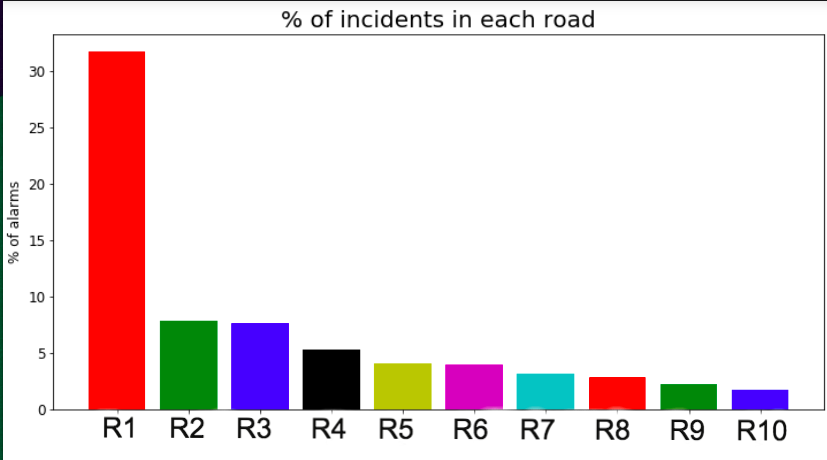} 
\caption{{\bf Most dangerous roads}}
\label{fig3}
\end{center}
\end{figure}

Figure 3 represents the percentage of the total amount of collision warnings on a road after normalization.

\begin{figure}[H]
\begin{center}
\includegraphics[width=85mm,height=50mm]{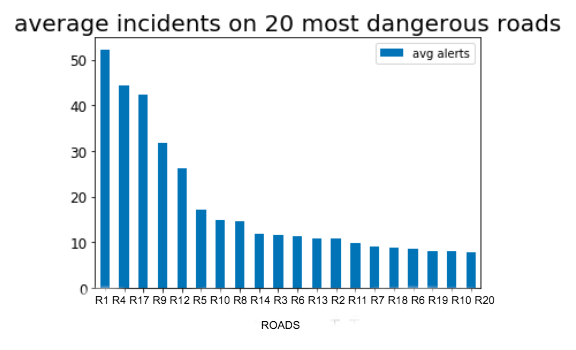} 
\caption{{\bf Average alerts on most dangerous roads}}
\label{fig4}
\end{center}
\end{figure}

Figure 4 represents the average number of collision warnings among different roads obtained from reverse geocoding using a Google Maps API.

\section{Predictive Modelling}
In the following sections, we use various Machine Learning approaches to model the complex interconnected input feature space- comprising the 14 dimensions of the TWL parameter list- to the output the safety index value between 1-5.
\subsection{Logistic Regression (LR)}
First, the problem statement was treated as a classification problem. In a binomial classification approach, the labels are assumed to be binary, indicating accident proneness (or not) using a 1 and 0. While an approach using the maximum likelihood estimation- where the dependent variable followed the Bernoulli Distribution- through a Logistic Regression model [10] is logically correct, it is not accurate due to the simplicity of the model.
\begin{equation}
 \frac{p}{1-p}=e^y
\end{equation}
where, 
 p is the probability of success
 \begin{equation}
 log(\frac{p}{1-p})=y
\end{equation}
\begin{equation}
 log(\frac{p}{1-p})=\beta_0 + \beta_1x
\end{equation}
When the safety index is considered as a label, ranging from 1-5, this becomes a multinomial classification problem, with the logistic regression model performing even more poorly. 
\subsection{Feed Forward Neural Network}
By increasing the complexity of the LR, with the use of hidden layers, the Feed Forward Neural Network (FFNN) was employed to model a safety index from a linear set of 14 independent inputs corresponding to the TWL specificities. However, this simple forward movement of information, adhering to delta rule and gradient descent methods, does not model the dependencies of the 14 interconnected vectors, with respect to its neighborhood [11].

\begin{equation}
{\displaystyle \Delta w_{ij}=-\eta {\frac {\partial E}{\partial w_{ij}}}=-\eta o_{i}\delta _{j}}\end{equation}
where
$w_{ij}$ is weight,
$\eta$ is the coefficient,
$E$ is the loss for the output,
$i,j$ are the neurons

The particular model used in this paper followed the backpropagation algorithm in accordance with Eqn. 5. The model was constructed using 2 Dense layers, each with 200 neurons each. The first 2 layers had an activation function of \textit{ReLu}, whereas the last layer had an activation function of \textit{Softmax}. The \textit{Adam} optimizer from Keras was used during compilation, while using \textit{Area Under Curve} as the accuracy metric.
\subsection{Long Short Term Term (LSTM) Neural Network}
To better the information persisting mechanism, the network was induced with loops, resulting in the LSTM approach. While this approach maintains information across a time series through a multiple copy method, it does not maintain the complexity of the relationship between the 14 parameters itself.
\begin{figure}[H]
\begin{center}
\includegraphics[width=70mm,height=40mm]{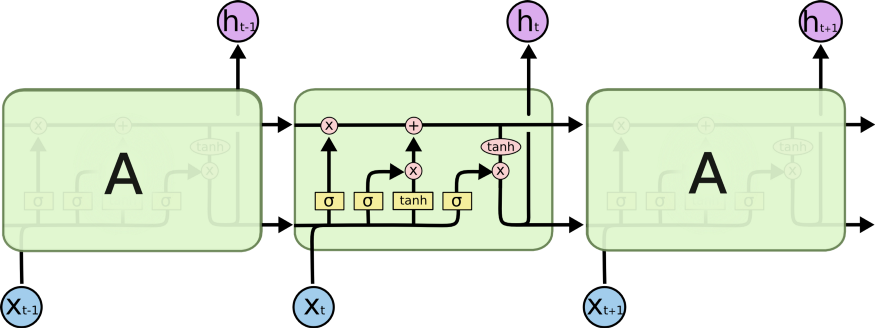} 
\caption{{\bf RNN Architecture}}
\label{fig5}
\end{center}
\end{figure}
The particular model used in this paper followed the \textit{current history} and \textit{previous history} formulae of RNNs. The model was constructed using 2 Dense layers, each with 200 neurons each. The first 2 recurrent layers had an activation function of \textit{ReLu}, whereas the last layer had an activation function of \textit{Softmax}. The \textit{Adam} optimizer from Keras was used during compilation, while using \textit{Area Under Curve} as the accuracy metric.
\subsection{Graph Neural Network}
To overcome the limitations of the previous models, the dependency between the 14-dimensional feature space and the associated accident proneness (or Safety Index) label has to be maintained across an arbitrary depth.

Thus, a Graph Neural Network (GNN) was used to employ a unique non-Euclidean data structure approach, using machine learning, which would focus on node classification, link prediction, and clustering. Graph data structures model a set of objects (nodes) and their relationships (edges). It works as a neighborhood aggregation scheme, where the node's representation vector is calculated by recursively aggregating and transforming the representation vectors of its neighboring nodes. 
 
The architecture of GNNs was first proposed in [12], which is an extension from the existing neural networks for processing the data represented in graph domains. In a graph, each node is defined by its features and
the related nodes. The objective of GNN is to learn the state
embedding $h_v \in R_s$ that contains the information and details of
the neighborhood for each node. The state embedding $h_v$ is an
s-dimension vector of node v and is used to produce
an output $o_v$ such as the node label. f denotes a parametric
function, called local transition function, which is shared among
all nodes and updates the node state according to the input
neighborhood. $g$ denotes the local output function that describes how the output is generated. After this, $hv$ and $ov$ are defined as follows:

\begin{equation}
h_v = f(x_v, x_{co[v]}, h_{ne[v]}, x_{ne[v]})
\end{equation}
\begin{equation}
o_v = g(h_v, x_v)\end{equation}
where $x_v, x_{co[v]}, h_{ne[v]}, x_{ne[v]}$ are the features of $v$, the features of its edges, the states, and the features of the nodes in
the neighborhood of $v$, respectively.
Let $H, O, X, and XN$ be the vectors constructed by
stacking all the states, all the outputs, all the features, and
all the node features, respectively. In a compact
form notation:
\begin{equation}
H = F(H, X)\end{equation}
\begin{equation}
O = G(H, X_N )
\end{equation}
where F, the $global transition function$, and G, the $global
output function$ are stacked versions of $f and g$ for all the nodes
in a graph, respectively. The value of $H$ is the fixed point of
Eq. 8 and is uniquely defined under the assumption that $F$ is
a contraction map.
Referencing Banach’s fixed point theorem [13],
the GNN uses the following classic iterative scheme for computing the state:

\begin{equation}
H^{t+1} = F(H^t, X)\end{equation}
where, $H^t$ denotes the $t-th$ iteration of H. The dynamical
system Eq. 9 converges exponentially fast to the solution, for any initial value H(0). 

The motivation for such a model stems from the comprehensive reviews on establishing a generalized method for the well-performing Convolutional Neural Networks (CNNs). CNN's extract multi-scale localized spatial features and compile them to build highly expressive representations, which
led to breakthroughs in almost all machine learning areas
and started the new era of deep learning [14]. The key takeaways from such an approach are the ideas of - local connection, shared weights and the use of multi-layer.[12] These are also of great importance in solving
problems of graph domain, because 1) graphs are the most
typical locally connected structure. 2) shared weights reduce
the computational cost compared with traditional spectral
graph theory [15]. 3) multi-layer structure is the key to deal
with hierarchical patterns, which captures the features of
various sizes. 
The limitation is that, CNNs can only operate on regular
Euclidean data like images and text- 2D grids and 1D sequences respectively- 
while these data structures can be regarded as instances of
graphs. Therefore, a generalization to such a CNN architecture is used, in the form of a GNN.

Another key aspect of such an approach would be the means of message passing. Graph embedding is an approach that is used to transform nodes, edges, and their features into vector space (a lower dimension) whilst maximally preserving properties like graph structure and information. Graphs are tricky because they can vary in terms of their scale, specificity, and subject.

Once the conversion of nodes and edges are completed, the graph performs Message Passing between the nodes. This process is also called Neighbourhood Aggregation because it involves pushing messages (embedding) from surrounding nodes around a given reference node, through the directed edges.

Sometimes, different neural networks are used for various types of edges; one for unidirectional and another for bidirectional. This way, the structure still capture the spatial relationship between nodes.
In the context of GNNs, for a single reference node, the neighboring nodes pass their messages/embedding through the edge neural networks into the recurrent unit on the reference node. The new embedding of the reference recurrent unit is updated by applying said recurrent function on the current embedding and a summation of the edge neural network outputs of the neighboring node embedding. Fig. 6 shows an illustration of this process where the blue square is a simple feed-forward NN applied on the embedding (white envelopes) from the neighboring nodes. The recurrent function (pink triangle) applied to the current embedding (white envelope) and summation of edge neural network outputs (black envelopes) to obtain the new embedding (white envelope prime).
\begin{figure}[H]
\begin{center}
\includegraphics[width=70mm,height=40mm]{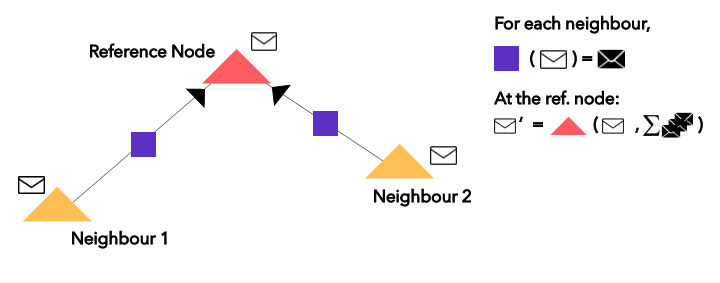} 
\caption{{\bf Message Passing via Neighbourhood aggregation}}
\label{fig6}
\end{center}
\end{figure}
This process is performed, in parallel, on all nodes in the network as embedding in layer L+1 depends on embedding in layer L. Which is why, in practice, there is no need to ‘move’ from one node to another to carry out Message Passing.
Also, the sum over the edge neural network outputs (black envelopes in the diagram) is invariant of the order of the outputs.
 
 The GNN model used in this particular paper follows the above-mentioned procedures of message passing, from layer to layer, while keeping the structure of the graph -in terms of input and output shape- similar. The unidirectional approach used here, follows a batch processing schema, where each batch consists of 20 graphs. Each graph pertains to a particular bus, traveling on a particular road, on a particular day. The individual nodes of this graph are the individual alert events that occurred for \textit{that} bus, on \textit{that} road, on \textit{that} day. Each event, in turn, comprises the 14-dimensional feature space, relating to the TWL parameters. The ordering of such events across the nodes is handled by the edges, which are time-series dependent and are thus directional in nature. The natural order being from the oldest event of the day, to the latest event of the day. Finally, a label- derived from the collision concentrations- is assigned to each graph, denoting the accident proneness on a scale of 1-5.  
 The architecture constructed is the first of its kind and is novel, as it employs a method of using a sequence of trainable graphs, each of which is attributed to a safety index. The identity- Session ID- is a string constructed by considering the device ID of the bus, the name of the road it traveled on, and that particular date (in standard format). Thus the Session IDs are unique to busses traveling on a given road, based on the dates.
 An Event is then defined using a vector squashing function. The squashing function takes into account the individual vectors which were considered as inputs to a model, as defined by the Principal Component Analysis. A total of 14 vectors, placed (logically) one above the other are finally squashed to a single 14-dimensional vector. This is treated as the Event Vector's feature space. Each event is a 14-dimensional feature space, corresponding to the respective TWL parameters, for a given time instant. 
 Next, the Session IDs are then made to correspond to a time-series ordered set of Events (that happened on that given day). Directed edges based on the time series of events, are then used to fit the structure onto a graph.
 
 Finally, a Label (binary or 1-5; based on the model) is attributed to each Session ID, which denotes the Accident Proneness or the Safety Index of the road respectively. This would serve analogous to the set of output labels present in the training data-set of any traditional Neural Network.
 
 This entire architecture, is used to train and model a GNN, to predict accident proneness and safety index of a road, based on the TWL feature space (14 in number).
 
 The architecture is represented in Figure 7.
 
 \begin{figure}[H]
\begin{center}
\includegraphics[width=92mm,height=60mm]{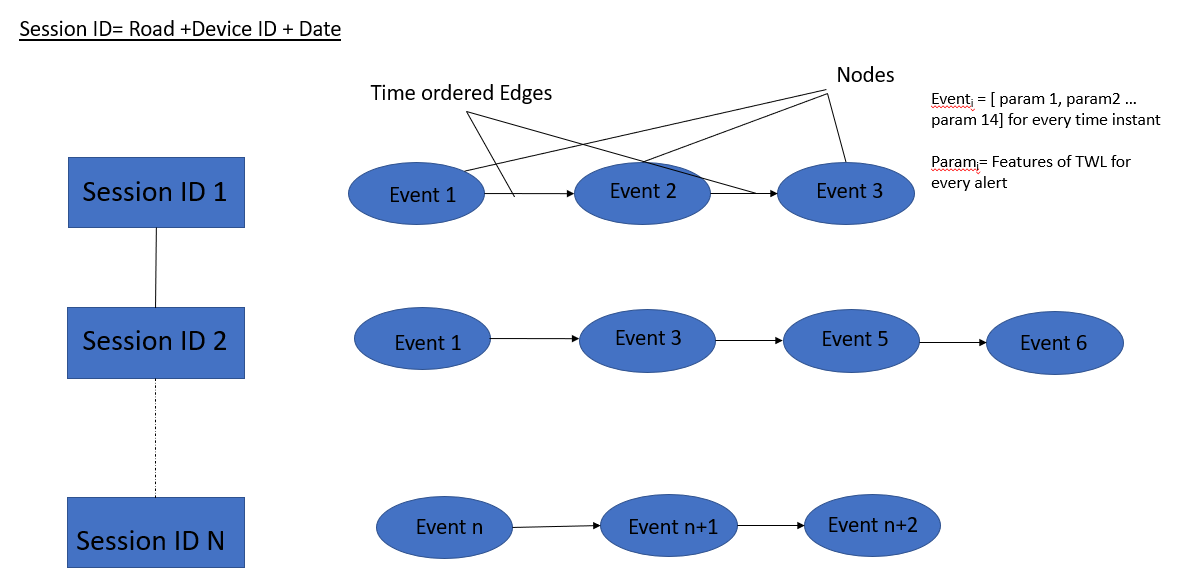} 
\caption{{\bf GNN Architecture}}
\label{fig7}
\end{center}
\end{figure}
 
 Individual batches are graph convoluted, and follows the graph message passing through a private self-implemented GraphConv (refer Pytorch) layer, termed SageConv.

\begin{equation}
\mathbf{x}_i^{(k)} = \gamma^{(k)} \left( \mathbf{x}_i^{(k-1)}, O_{j \in \mathcal{N}(i)} \, \phi^{(k)}\left(\mathbf{x}_i^{(k-1)}, \mathbf{x}_j^{(k-1)},\mathbf{e}_{j,i}\right) \right),
 \end{equation}
 
 where x is node embedding, e is edge feature,
 
 $\phi$
 is message function,
 $O$
 is aggregation function,
 
 $\gamma$
 is an update function.
  
\section{Results}
After normalizing the data, a PCA and an exploratory study identified 14 TWL features that influence the safety of a road. These features were used to train various machine learning models, explained in the previous section, to be able to predict the safety index value of a new location.

\begin{figure}[H]
\begin{center}
\includegraphics[width=70mm,height=40mm]{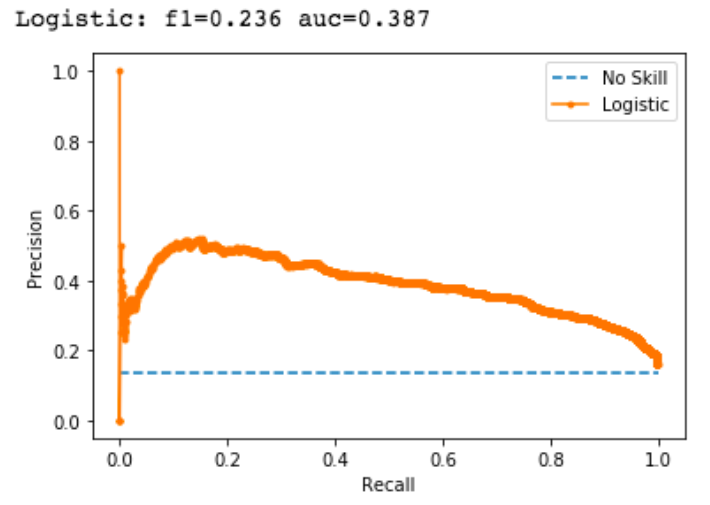} 
\caption{{\bf Logistic Regression}}
\label{fig8}
\end{center}
\end{figure}

Fig. 8 shows the result of using logistic regression. An accuracy of 38.7\% is achieved for the test data after training the model.

\begin{figure}[H]
\begin{center}
\includegraphics[width=70mm,height=40mm]{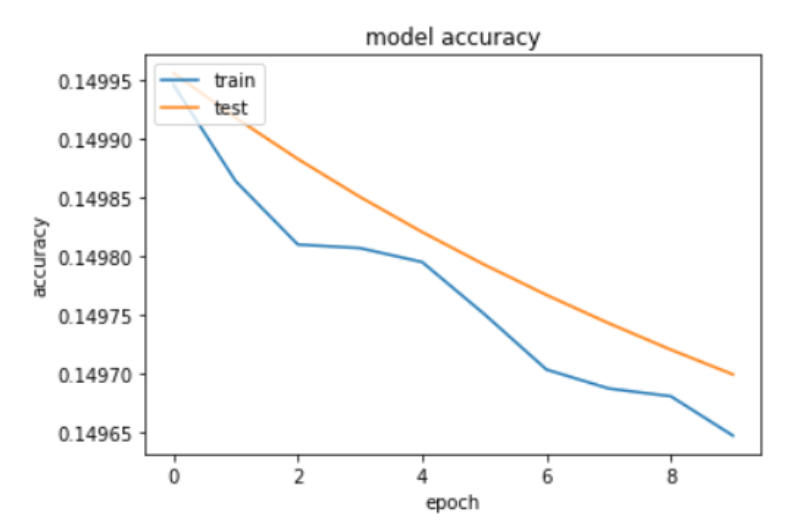} 
\caption{{\bf Simple Feed Forward Neural Network}}
\label{fig9}
\end{center}
\end{figure}

Fig. 9 shows the result of using a simple feed-forward neural network. This method returns poor results, converging to an accuracy of below 14.93\%

\begin{figure}[H]
\begin{center}
\includegraphics[width=70mm,height=40mm]{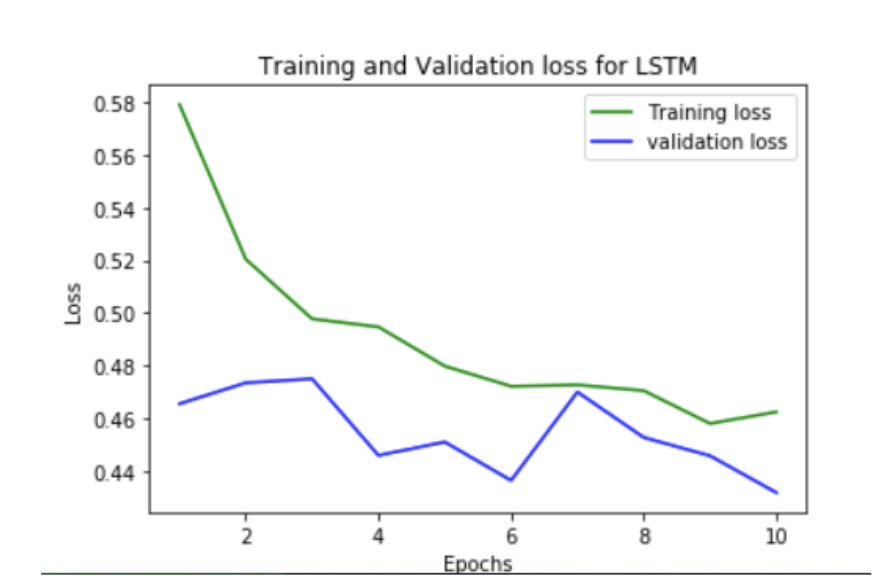} 
\caption{{\bf Recurrent Neural Network (LSTM)}}
\label{fig10}
\end{center}
\end{figure}

Fig. 10 displays the result of using a type of recurrent neural network (LSTM). This method shows better results than logistic regression due to its ability to preserve and utilize historical data. It converges to an accuracy of 42.6\%.

\begin{table*}
\centering
Table \ref{table:1} shows the Epochs, Loss and the AUC Accuracies for the GNN model.
\begin{center}
\begin{tabular}{ |c|c|c|c|c| } 
 \hline
 Epoch:053, & Loss:0.03829, & Train Auc: 0.99833, & Val Auc: 0.61129, & Test Auc:0.65997 \\ 
 Epoch:054, & Loss:0.03528, & Train Auc: 0.99865, & Val Auc: 0.61163, & Test Auc:0.66098 \\ 
 Epoch:055, & Loss:0.02899, & Train Auc: 0.99879, & Val Auc: 0.60948, & Test Auc:0.65792 \\ 
 Epoch:056, & Loss:0.02696, & Train Auc: 0.99886, & Val Auc: 0.60899, & Test Auc:0.65772\\
 Epoch:057, & Loss:0.02577, & Train Auc: 0.99880, & Val Auc: 0.61097, & Test Auc:0.66086\\
 Epoch:058, & Loss:0.02742, & Train Auc: 0.99863, & Val Auc: 0.61275, & Test Auc:0.66381\\
 Epoch:059, & Loss:0.02781, & Train Auc: 0.99887, & Val Auc: 0.61476, & Test Auc:0.66355\\
 Epoch:060, & Loss:0.02481, & Train Auc: 0.99895, & Val Auc: 0.61189, & Test Auc:0.66245\\
 \textbf{Epoch:061,} & \textbf{Loss:0.02434,} & \textbf{Train Auc: 0.99890,} & \textbf{Val Auc: 0.61377,} & \textbf{Test Auc:0.65856}\\
 \hline
\end{tabular}
\end{center}
\caption{Graph Neural Network}
\label{table:1}
\end{table*}

\bigskip
Table \ref{table:1} describes the output log of the Graph Neural Network. The training result is shown to converge after approximately 60 epochs. A significantly higher accuracy of 65\% is achieved using a GNN. The difference in performance can be attributed to the GNN's inherent ability to link and process information in a higher dimensional feature space than the previously tested methods. Hence the GNN was experimentally validated to produce better results for our application.
The modelling works to comprehend a complex network of information between the 14 dimensions of the TWL feataure space, and predict an associated label from 1 to 5- which denotes the safety index. As is seen through the accuracy log via the area under curve metric, the theoretical results are validated. While it does not directly speak about the generalization ability of GNNs, it is reasonable to notice that GNNs with strong expressive power can accurately capture such graph structures of interest and model them well. This shows that treating the feature space and respective alerts, as a graph with a time series ordered set of events as nodes- across a particular road on a specific days for every bus- is a valid approach to solving this problem statement.

\section{Conclusion}
 The prediction of road accidents is associated with a high degree of uncertainty due to human error accounting for a major fraction of the causes.
 In this paper, the accident proneness of a road caused by environmental and non-human dynamic factors has been predicted. The concentrations of collision warnings, after normalization and development into a "Safety Index", was used as an indication as to the vulnerability of a location to road accidents. The influence features associated with time, weather, and location was quantified based on the effect they had on the Safety Index and Accident Proneness. Multiple machine learning models were tested on the data to predict the Safety Index of a location-based on TWL characteristics. The Graph Neural Network approach achieved the best results with an accuracy of 65\% on untrained data. The reason for not achieving an even higher accuracy, can be attributed to the randomness associated with human error in many accidents. Future models could also use a different algorithm on fitting the connections of this graph structure, to try yielding better results. One such approach could be by setting the connections and ordering of events (nodes) of the graph - instead of just following a time series order- to follow the 3D distance formula by including the latitude and longitude along with time, as parameters to decide the next node. 
\section*{ACKNOWLEDGEMENT}
This work was done in part under the Intel Corporation Student Contest 2020 in association with the Center for Innovation and Entrepreneurship (CIE) at PES University. We acknowledge all the support from Intel Corporation towards this work. For this, we would like to thank the CIE of PES University and our mentors from Intel Corporation for providing us with this opportunity and for their continuous support. We extend our thanks to PES University, who provided us a platform that helped us to team up and pursue this project. We also thank Dr. Asif Qamar, for his help in giving us certain key insights. Special thanks to Prof. Madhukar Narasimha and Prof. Sathya Prasad for their constant guidance.
 


\end{document}